\documentclass[10pt,twocolumn,letterpaper]{article}

\usepackage{wacv}              

\usepackage{graphicx}
\usepackage{amsmath}
\usepackage{amssymb}
\usepackage{booktabs}
\usepackage{algpseudocode}  
\usepackage{algorithm}
\usepackage{siunitx}
\usepackage{tabularx}
\usepackage{multirow}
\usepackage{setspace}
\usepackage{hyperref}
\usepackage[accsupp]{axessibility}

\algnewcommand{\IState}[1]{\Statex \hspace{-\algorithmicindent} #1}

\usepackage[capitalize]{cleveref}
\crefname{section}{Sec.}{Secs.}
\Crefname{section}{Section}{Sections}
\Crefname{table}{Table}{Tables}
\crefname{table}{Tab.}{Tabs.}

\def\wacvPaperID{1620} 
\def\confName{WACV}
\def\confYear{2024}

\begin{document}

\title{Self-Sampling Meta SAM: Enhancing Few-shot Medical Image Segmentation with Meta-Learning}

\author{Tianang Leng{\thanks{equal contribution} }\\
Huazhong University of Science and Technology\\
Wuhan, China\\
{\tt\small tianangl@hust.edu.cn}
\and
Yiming Zhang{\footnotemark[1]}\\
Tokyo Institute of Technology\\
Tokyo, Japan\\
{\tt\small zhang.y.bl@m.titech.ac.jp}
\and
Kun Han\\
University of California, Irvine\\
Irvine, California, United States\\
{\tt\small khan7@uci.edu}
\and
Xiaohui Xie\\
University of California, Irvine\\
Irvine, California, United States\\
{\tt\small  xhx@ics.uci.edu}
}

\maketitle

\begin{abstract}
While the Segment Anything Model (SAM) excels in semantic segmentation for general-purpose images, its performance significantly deteriorates when applied to medical images, primarily attributable to insufficient representation of medical images in its training dataset. Nonetheless, gathering comprehensive datasets and training models that are universally applicable is particularly challenging due to the long-tail problem common in medical images.

To address this gap, here we present a \textbf{\underline{S}elf-\underline{S}ampling \underline{M}eta \underline{SAM}} (SSM-SAM) framework for few-shot medical image segmentation. Our innovation lies in the design of three key modules: 1) An online fast gradient descent optimizer, further optimized by a meta-learner, which ensures swift and robust adaptation to new tasks. 
2) A Self-Sampling module designed to provide well-aligned visual prompts for improved attention allocation; and 3) A robust attention-based decoder specifically designed for few-shot medical image segmentation to capture relationship between different slices.
%
Extensive experiments on a popular abdominal CT dataset and an MRI dataset demonstrate that the proposed method achieves significant improvements over state-of-the-art methods in few-shot segmentation, with an average improvements of 10.21\% and 1.80\% in terms of DSC, respectively. 
In conclusion, we present a novel approach for rapid online adaptation in interactive image segmentation, 
adapting to a new organ in just 0.83 minutes. Code is available at \href{https://github.com/DragonDescentZerotsu/SSM-SAM}{https://github.com/DragonDescentZerotsu/SSM-SAM} 
\end{abstract}

\section{Introduction}

Medical image segmentation plays a pivotal role in clinical applications, including disease diagnosis, abnormality detection and treatment planning. Historically, the segmentation of anatomical structures was performed manually by experienced physicians, a laborious and time-consuming task.

Recent advancements in deep learning offer automated segmentation tools that deliver near-human accuracy swiftly. Nevertheless, these tools need extensive training on large annotated datasets for optimal performance, which are costly and time-intensive since they require inputs from experts with extensive clinical experience. Thus, in the field of medical imaging, few-shot learning \cite{snell2017prototypical, li2021adaptive, zhang2021self} has gained significant interest from researchers because it is able to segment accurately without extensive labeled data. In fact, to achieve optimal performance on unseen classes, few-shot learning models must excel in extracting representative features from limited data. However, current few-shot learning frameworks for medical image segmentation \cite{dong2018few, wang2019panet, tang2021recurrent} predominantly pretrain their models using data from a single medical domain, often with restricted datasets, leading to limited feature extraction ability. If we can leverage the strong feature extraction ablities obtained from extensive training data of diverse domains, the model can capture more distinctive features, enhancing its adaptability to new tasks. 

Recently, Segment anything model (SAM) \cite{kirillov2023segment} has attracted significant attention. Trained on a large segmentation dataset of over 1 billion masks across various domains of natural images, SAM has strong feature extraction and generalization abilities and can segment any object on a certain image. Given SAM's excellent zero-shot transferability, a natural idea is to directly apply SAM for medical image segmentation to address the issue of relatively scarce medical image data \cite{roy2023sam, he2023accuracy}. However, recent studies \cite{mazurowski2023segment, shi2023generalist} have shown that the performance of SAM is overall moderate and varies significantly across different datasets and different cases, demonstrating the potential promise of SAM within the context of medical images but also shows that the model cannot be applied directly  with high confidence. 
Such phenomenon is attributed to the vast differences between natural and medical images, as SAM was primarily trained on natural images.
To better adapt SAM to medical images, most prior studies focused on integrating lightweight Adapters \cite{chen2023sam, wu2023medical} or on freezing the heavy image and prompt encoders, opting to fine-tune solely the mask decoder \cite{ma2023segment, hu2023skinsam, li2023polyp}. However, the limited data available for unseen classes will hinder the few-shot segmentation performance of these approaches.

Driven by the desire to maximize the powerful extracted features of SAM through prompts \cite{zhang2023survey}, and to leverage its potential zero-shot capabilities without vast training data, we introduce the Self-Sampling Meta SAM (SSM-SAM) framework, which utilizes a Meta-learning method MAML++. Our framework consists of two main parts. The first part is a redesigned backbone that removes the original SAM's prompt encoder and mask decoder; instead, we employ a self-sampling prompt encoder and Flexible Mask Attention Decoder (FMAD). Also, we integrate adapters into the image encoder as previous works do, allowing SAM to learn features of unfamiliar tasks. This enables SAM to have better transferability for few-shot learning and is termed \textbf{SS-SAM} (\underline{S}elf-\underline{S}ampling \underline{SAM} without MAML++). The second part, which we refer to as \textbf{SSM-SAM} (with MAML++), serves as our final framework for few-shot learning. It is a meta-learning based optimizer layered on top of this backbone to further boost SAM's few-shot performance on medical images.
 
We performed experiments for few-shot learning on an abdomen CT dataset and an MRI dataset. We also utilized a fully supervised medical image segmentation task to evaluate the performance of our SS-SAM backbone (without MAML++) in comparison to prior methods on CT dataset. 

Our contributions are summarized as follows:
\begin{itemize}
    \item An effective online parameter adaptation technique  optimized by a MAML++ \cite{antoniou2018train} based meta-learner to enhance SAM's generalization and few-shot learning capacities. 
    \item A positive-negative self-sampling module that can generate aligned visual prompts to better extract the contextual relationship.
    \item A novel Flexible Mask Attention Decoder specifically designed for medical image few-shot segmentation.
    \item Our method outperformed the SOTA framework for few-shot medical image segmentation by an average of 10.21\% on MICCAI15 Multi-Atlas Abdomen Labeling challenge dataset \cite{MICCAI} and 1.80\% on ISBI 2019 Combined Healthy Abdominal Organ Segmentation Challenge \cite{kavur2021chaos} in terms of DSC.
\end{itemize}

Each module we use is plug-and-play, aiming to facilitate the deployment of fast and robust online image segmentation systems across various industries and making our model a baseline for improving performance where foundational models like SAM have struggled in the past.

\section{Related Work}
\begin{figure*}[ht]
\centering
\includegraphics[height=4cm]{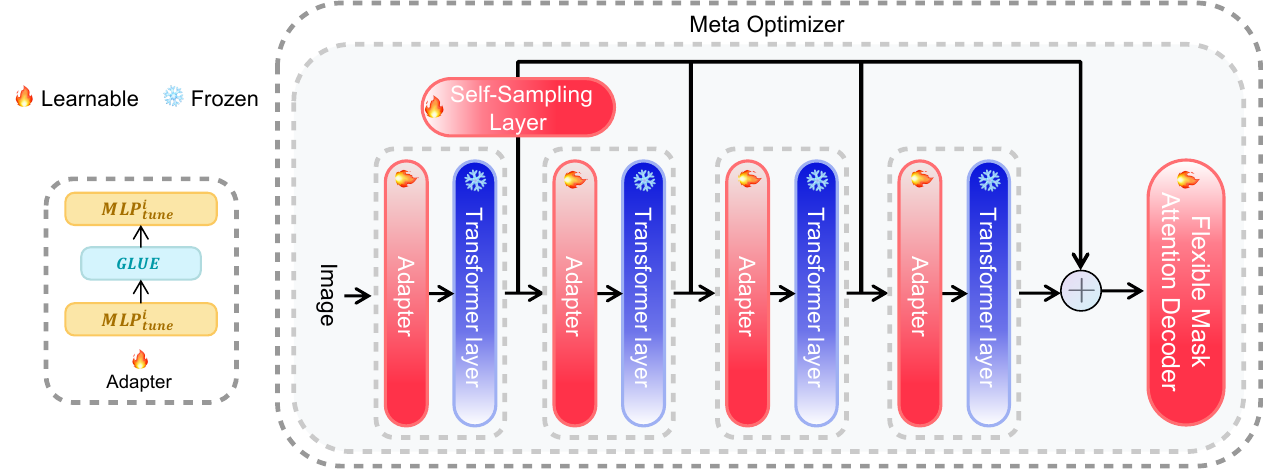}
\caption{Overview of our proposed framework for SSM-SAM. SSM-SAM backbone consists of three main components:
a) Injecting Adapters into the encoder of SAM to quickly acquire task-related information.
b) Using the Self-Sampling Module to replace the position encoding based prompt encoder to strengthen the relationship between the feature and the prompt.
c) Flexible Mask Attention Decoder (FMAD) to enhance boundary information and refine the final generated mask map. On top of the backbone, we employ a meta-learning based online optimizer.}
\label{fig:1}
\end{figure*}

\noindent\textbf{Foundation Models.} A foundation model is essentially a large pre-trained model, typically developed using self-supervised learning across diverse datasets. This allows it to be fast adapted to specific tasks through mechanisms by fine-tuning or in-context learning. At present, foundation models have reached a level of maturity in the NLP domain, with models like BERT \cite{devlin2018bert}, GPT-3 \cite{brown2020language}. 
 In the realm of computer vision, the Segment Anything Model (SAM) \cite{kirillov2023segment}, pre-trained on 11M images with 1B masks, first introduced a foundation model for image segmentation. This model can interpret a variety of prompts given by users, such as points, boxes, masks, and texts, to generate user-specified segmentations and is adept at segmenting any object within an image. However, despite its zero-shot generalization capabilities, SAM has demonstrated sub-optimal performance across multiple downstream tasks \cite{ji2023segment}. This observation has stimulated ongoing research efforts to enhance SAM's performance in these tasks \cite{ma2023segment, wu2023medical, gong20233dsamadapter, zhou2023sam}. \\
\noindent\textbf{Adapters.} 
Adapters have emerged as a powerful tool in the realm of transfer learning for NLP and computer vision. Introduced as a lightweight alternative to full model fine-tuning, adapters allow for model customization while preserving the pre-trained parameters \cite{houlsby2019parameter}. Instead of updating all the parameters in the model, adapter methods insert small bottleneck layers in the model architecture, which are then fine-tuned while the original model parameters remain unchanged. Recently, the ViT-Adapter \cite{chen2022vision} has been developed to equip a standard ViT \cite{dosovitskiy2021image} with the capability to perform a variety of downstream tasks. In a further development, an Explicit Visual Prompting (EVP) \cite{liu2023explicit} technique has facilitated the integration of explicit visual cues into the Adapters. In this work,  We integrate such Adapters \cite{chen2023sam} into the image encoder of SAM to avoid training a large amount of parameters. \\
\noindent\textbf{Visual Prompts.} 
The application of visual prompts in computer vision tasks has its roots in interactive segmentation, a methodology that necessitates user input, such as clicks \cite{xu2016deep, wang2018interactive, jang2019interactive, lin2020interactive, chen2021conditional}, bounding boxes \cite{wu2014milcut, rajchl2016deepcut}, or scribbles \cite{batra2010icoseg, bai2014error, lin2016scribblesup}, or convert spatial queries to masks and feed them into the image backbone \cite{liu2022simpleclick} to assist the algorithm in accurately delineating object boundaries. These visual prompts augment the segmentation process, yielding more accurate and dependable outcomes \cite{wang2023review}. 
However, such form of visual prompts either struggle to adapt to unseen prompts \cite{kirillov2023segment} or might be heavy in applications as each interaction necessitates processing the image through the feature extractor \cite{liu2022simpleclick}. Recently, visual sampler has been introduced to transform all non-textual queries into visual prompts that reside within the same visual embedding space to address these limitations \cite{zou2023segment}. However, no prior research has endeavored to employ this approach for generating prompts using the SAM trained on an extensive image corpus. Here we seek to bridge this research gap. \\
\noindent\textbf{Model Agnostic Meta Learning (MAML).} One continuous challenge of few-shot medical image segmentation is the distribution mismatch between training and testing datasets, particularly the long tail problem. 
Fortunately, MAML \cite{finn2017model} and MAML++ \cite{antoniou2018train} offers meta-learning frameworks tailored to combat this issue. It is elegantly simple yet can find suitable model-agnostic initialization parameters that are trained through various tasks and can quickly adapt to new tasks. Many previous works like\cite{makarevich2021metamedseg,singh2021metamed,yu2020foal} use meta-learning to address different problems in medical image segmentation. Yet, MAML++ framework has been primarily focused on vision tasks like classification and recognition. All these previous works inspired us to utilize MAML++ to enhance foundational models' few-shot learning ability.

\section{Methodology}
\subsection{Overview of SSM-SAM architecture}
\cref{fig:1} depicts the architecture of our few-shot learning framework.
We keep the image encoder of SAM frozen and add a learnable \textbf{Adapter} layer \cite{liu2023explicit} to each of the transformer layers for parameter-efficient training. After passing through the image encoder, we perform a self-sampling operation on the first image embedding to update it. Afterward, we feed the four output image embeddings into our Flexible Mask Attention Decoder (FMAD) 
to refine these embeddings and generate the predicted mask. On top of this backbone, we implement a MAML++ based meta-learner to search for optimal initialization parameters for rapid adaptation to different organs.

\subsection{Few-shot Online Optimizer}
Segmentation models' clinical deployment has historically been hindered by a distribution mismatch between training and testing datasets. Given the impossibility of collecting comprehensive representative data, we have innovated in our training approach. Rather than creating a universal offline model, we designed our model to recognize and adapt to new data types, ensuring it remains relevant for unseen images. In our few-shot medical image segmentation scenario, we consider a distribution over organs, denoted as $p(O)$, to which we aim our model to adapt. We represent our model as $f_\theta$ with parameter $\theta$, which is trained on images of different organs, $I_{o_i}$, following the distribution $p(I_{o_i})$, where $o_i$ signifies the $i^{th}$ organ. The online optimizer optimizes the parameters via back-propagating steps as follows:
\begin{equation}
    \theta_i^{'} \leftarrow \theta_i^{'}-\alpha\nabla_{\theta_i^{'}}L_{o_i}(f_{\theta_i^{'}})
\end{equation}
where $\alpha$ is the learning rate and $\theta_i^{'}$ represents the model parameters adapted to fit the $i^{th}$ organ. We employ balanced cross-entropy and IoU loss to supervise our network so $L_{o_i}$ can be expressed as: 
\begin{equation}
\label{loss}
    L_{o_i} = L_{bce} + L_{iou}
\end{equation}
The overview of the online optimizer is outlined in \cref{online}.

\begin{algorithm}
\caption{SSM-SAM Online Optimizer}
\label{online}
\begin{algorithmic}[1] 
\IState \textbf{Input:} $K$ image-mask pairs $P_{i}=\{(I_{i},M_{i})_k\}$ of the target organ $o_i$
\IState \textbf{Input:} trained parameters: $\theta$
\IState \textbf{Require:} learning rate $\alpha$; number of online optimization steps $S$
\State $\theta_i^{'} \gets \theta$
\For{$s$ from 1 to $S$}
    \For{$(I_i,M_i)_k$ in $P_i$}
        \State Update parameters with gradient descent:
        \Statex \hspace{3em} $\theta_i^{'} \gets \theta_i^{'}-\alpha\nabla_{\theta_i^{'}}L_{O_i}(f_{\theta_i^{'}}((I_i,M_i)_k))$
    \EndFor
\EndFor
\IState \textbf{Output:} updated parameters: $\theta_i^{'}$
\end{algorithmic}
\end{algorithm}

\subsection{Meta-learning}
We employ meta-learning for our model training to guarantee that the initial parameters $\theta$ exhibit robust generalization capabilities, facilitating rapid online optimization. MAML++ \cite{antoniou2018train} is an ideal strategy for our purpose, as it is designed to learn suitable initial model parameters optimized for swift adaptation to new tasks. The complete algorithm is detailed in Algorithm \ref{meta-learning}.
\begin{algorithm}
\caption{SSM-SAM Offline Meta-learner}
\label{meta-learning}
\begin{algorithmic}[1] 
\IState \textbf{Input:} organ set $O$; initial weight and learning rate $\theta,\beta$
\IState \textbf{Require:} number of epochs $E$; number of within organ optimization steps $S$; online learning rate $\alpha$
\For{$e$ from 1 to $E$}
    \State Sample $T$ organs $\{o_1,o_2,...,o_T\}$ from $O$
    \For{$i$ from 1 to $T$}
        \State $\theta_i^{'} \gets \theta$
        \State Sample $K$ image-mask pairs $P_{i}=\{(I_{i},M_{i})_k\}$ 
        \Statex \hspace{3em}of the target organ $o_i$
        \State $\theta_i^{'}\gets$\textbf{Online Optimizer}($S,P_i,\theta_i^{'},\alpha$)
    \EndFor
    \State Resample $K$ images-mask pairs $P_{i}^{'}$ for each $o_i$
    \State $\beta \gets \textbf{CosineAnnealingLR}(\beta,e)$
    \State $\theta \gets \theta-\beta\nabla_{\theta}\frac{1}{T}\sum_i^{T}L_{o_i}(f_{\theta_i^{'}}(P_i^{'}))$
\EndFor
\IState \textbf{Output:} updated parameters: $\theta$
\end{algorithmic}
\end{algorithm}
Note that 
 the outermost for-loop is the meta-learner, which is defined as follows:
\begin{equation}
    \theta \gets \theta-\beta\nabla_{\theta}\frac{1}{T}\sum_i^{T}L_{o_i}(f_{\theta_i^{'}})
\end{equation}
where $i$ is the $i^{th}$ organ. $T$ is the number of organs in a task batch for optimizing the meta-learner. We incorporate the cosine annealing learning rate as recommended in \cite{antoniou2018train} to enhance our model's efficiency in fitting the training set. 

\subsection{Adapted Image Encoder}
The image encoder of SAM uses a Vision Transformer (ViT) \cite{dosovitskiy2021image}, which is pre-trained with MAE \cite{he2022masked}. When given an image of any size, it is essential to first resize it to a resolution of 1024 x 1024. The ViT then processes this resized image to generate an image embedding with dimensions $C\times H\times W$. For this research, we chose the ViT-B variant (SAM-b), (where $C$ = 256, $H$ = 64, $W$ = 64).  To enable efficient learning with faster updates and address the issue of excessive GPU memory usage, we keep the image encoder frozen and inject a trainable \textbf{Adapter} to each of the 12 transformer layers as mentioned before. We only train the parameters within \textbf{Adapters}, which are tasked with learning task-specific knowledge and low-level structural information from the features extracted from image embeddings. For the $k$-$th$ Adapter layer, we take patch embedding $\text{F}^{k}$ as input and obtain updated parameters $\text{P}^{k}$
\begin{equation} 
\text{P}^{k} = \text{{MLP}}_{\text{{up}}}(\text{{GELU}}(\text{{MLP}}^{k}_{\text{{tune}}} (F^{k}))
\end{equation}
where $\text{{MLP}}^{k}_{\text{{tune}}}$ refers to a linear layer within each Adapter for generating distinct prompts. $\text{{MLP}}_{\text{{up}}}$ is an up-projection layer shared across all Adapters designed to match the dimension of transformer features. $\text{P}^{k}$ represents the output associated with each transformer layer. 

Also, we divided image encoder to four sub-blocks, which is carried out with the intent of deriving multi-level information \cite{li2022exploring}. After passing through the image encoder, these four image embeddings of size $B \times 256 \times 64 \times 64 $ are fed into the subsequent module.

\subsection{Self-Sampling Module}

\begin{figure}[ht!]
\centering
\includegraphics[height=5cm]{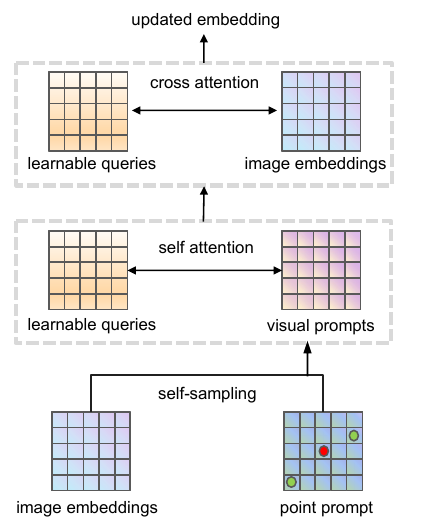}
\caption{Structure of queries and prompt interaction during training and evaluation.}
\label{fig:2}
\end{figure}

\begin{figure*}[ht!]
\centering
\includegraphics[height=7cm]{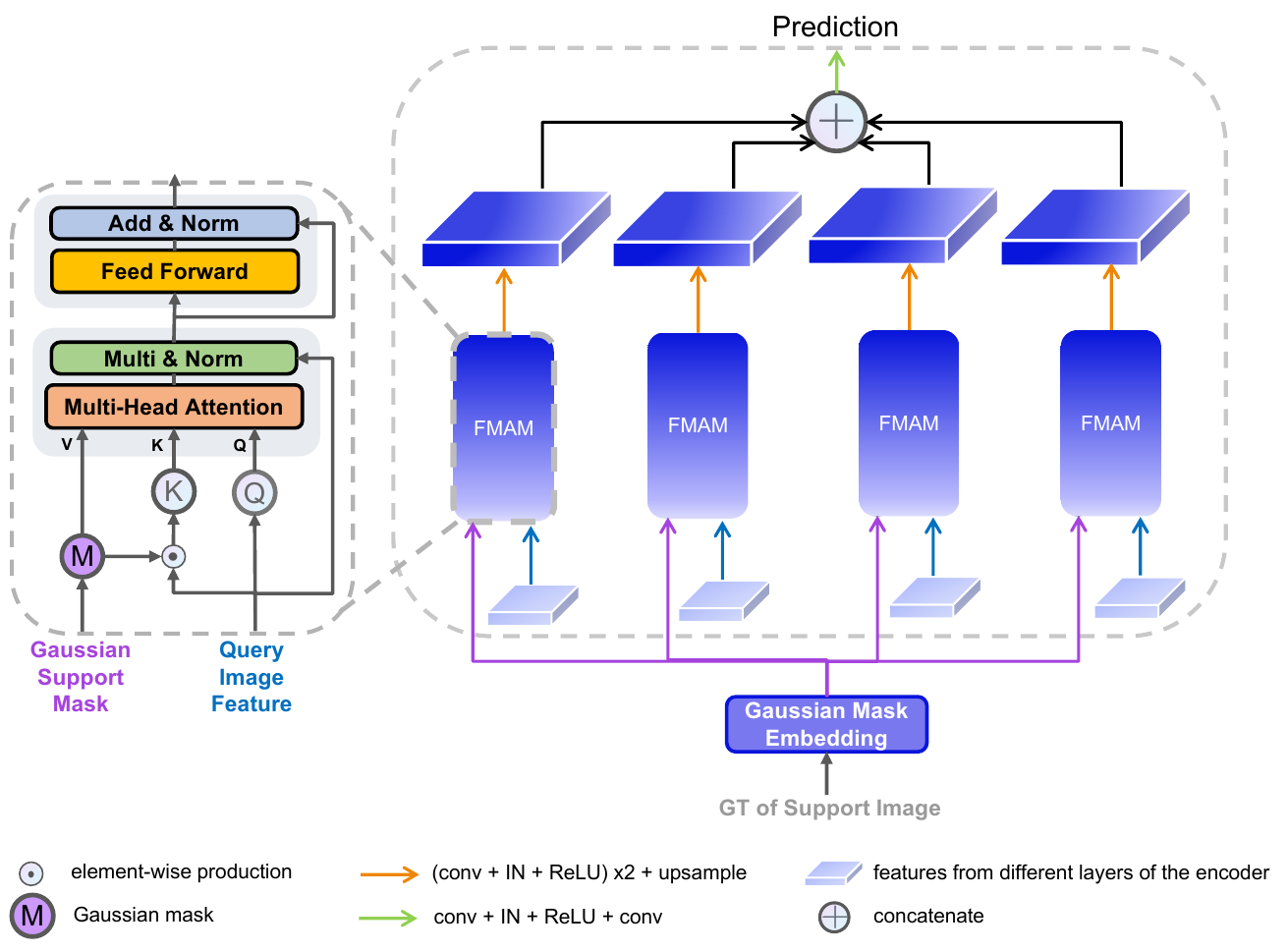}
\caption{Process of Flexible Mask Attention Module (FMAM) and Flexible Mask Attention Decoder(FMAD).}
\label{FMAM}
\end{figure*}
We employ a \textit{positive-negative attention self-sampling} module inspired by \cite{zou2023segment} to convert all kinds of non-textual queries to visual prompts that lie in the same visual embedding space, as opposed to the conventional position encoding in SAM\cite{kirillov2023segment}. Our algorithm is illustrated in \cref{fig:2} and can be summarized as follows:
\begin{gather}
    P = \textbf{concatenate}([P_{postive}, P_{negative}]) \\
    P_{v} = \textbf{Self-Sampler} (P, Z_{h})
\end{gather}
where $P_{positive}$, $P_{negative}$ stand for positive points (region we aim to segment) and negative points (background region) sampled from the image, respectively. $Z_{h}$ is the feature maps extracted from the image and $P$ denotes the point prompts. 

Afterward, we concatenated these positive points and negative points to form the point prompts. This approach ensures that the model does not only focus on the positive parts, which could potentially lead to a higher false positive rate, but also adequately attends to the negative parts. With this method, the model can better discerns the organ boundaries, enhancing segmentation performance when we only give one point to prompt the model during inference.

On receiving the point prompts, we perform direct interpolation from the feature map to get the corresponding embeddings as visual prompts. This ensures that the prompts and the image embeddings share identical semantic features. We initialize a few tokens as learnable global queries and concatenate them with these visual prompts. After that, we apply self-attention \cite{vaswani2017attention} between a set of learnable queries and the visual prompts. Then, cross-attention is applied only between the image embeddings and these updated queries. 
 
\subsection{Flexible Mask Attention Decoder}
Since we divided each 3D volume data into 12 chunks as in section \ref{setup}, there is a certain pattern to how each organ changes across different slices. This observation inspired us to treat the segmentation task like a tracking task. Consequently, we introduced the Flexible Mask Attention Decoder (FMAD). The core of this decoder is the Flexible Mask Attention Module (FMAM), as illustrated in Fig \ref{FMAM}. By leveraging cross-attention, we can expand or contract the blurred support mask to predict the mask for the query image, mitigating the challenge of direct mask prediction. The Gaussian blurred support masks provide the model with ample room to explore potential mask positions while efficiently suppressing distractions from other parts of the image. 

In decoding, the attention block bridges the Gaussian blurred support mask and queries images to update the mask. We first apply a Gaussian filter on original support mask $\textbf{M}\in\mathbb{R}^{H\times W\times 1}$, repeat it for $C$ times and reshape to $N\times C$ to get $\textbf{M}^{'}\in\mathbb{R}^{N\times C}$. Then we can easily cast it on query feature $\textbf{F}\in \mathbb{R}^{N\times C}$ extracted from image encoder to get $\textbf{K}\in \mathbb{R}^{N\times C}$ and $\textbf{Q}\in \mathbb{R}^{N\times C}$ as follows:
\begin{equation}
    \textbf{K}=(\textbf{M}^{'}\odot \textbf{F})\textbf{W}^{K} \\
\end{equation}

where $\odot$ means element-wise production, $\textbf{W}^{K}\in\mathbb{R}^{C\times C}$ means key projection matric. Following \cite{vaswani2017attention}, we also adopt the dot-product to compute the similarity matrix $\textbf{A}_{K\to Q}\in\mathbb{R}^{N\times C}$ between the query and key as follows:
\begin{equation}
    \textbf{A}_{K\to Q}=\mathrm{Atten}(\textbf{Q},\textbf{K})=\mathrm{Softmax_{col}}(\bar{\textbf{Q}}\bar{\textbf{K}}^{T}/\tau)
\end{equation}

where $\bar{\textbf{Q}}$ and $\bar{\textbf{K}}$ are $l_2$-normalized features of \textbf{Q} and \textbf{K} across the channel dimension, and $\tau$ is a temperature parameter controlling the Softmax distribution, which is the same as in \cite{wang2021transformer}. Then we convert $\textbf{M}^{'}$ to $\textbf{V}\in\mathbb{R}^{N\times C}$ though:
\begin{equation}
    \textbf{V}=(\textbf{M}^{'})\textbf{W}^{V}
\end{equation}

where $\textbf{W}^{V}\in\mathbb{R}^{C\times C}$ denotes value projection matrix. With the attention matrix $\textbf{A}_{K\to Q}$ from key to query, we can transform the value via $\textbf{A}_{K\to Q}\textbf{V}\in \mathbb{R}^{N\times C}$.

Unlike typical attention layers, at the first residual connection, we use \textit{multiply} \& \textit{Norm} instead of \textit{Add} \& \textit{Norm} to propagate query feature \textbf{F} as follows:
\begin{equation}
    \textbf{F}^{'}=\mathrm{InsNorm}(\textbf{A}_{K\to Q}\textbf{V}\odot\textbf{F})
\end{equation}
By using the updated Gaussian mask from the attention layer, we can further refine the predicted mask. After processing each of the four image embeddings, we upsample, concatenate, and put them through a convolutional layer. This approach facilitates improved information integration across layers to predict the final mask.

\section{Experiments}
\subsection{Setup}
\label{setup}
\textbf{Datasets} To verify the generality and robustness of our approach, we perform experiments on two datasets, ABD-30 and ABD-MRI. We evaluate the complete few-shot segmentation framework (SSM-SAM) on both datasets, while the backbone (SS-SAM) is only assessed on ABD-30 to validate its efficiency.

- ABD-30 (from MICCAI15 Multi-Atlas Abdomen Labeling challenge\cite{MICCAI}) contains 30 cases with 3779 axial abdominal clinical CT images. 

- ABD-MRI (from Combined Healthy Abdominal Organ Segmentation (CHAOS) challenge \cite{kavur2021chaos} held in IEEE International Symposium on Biomedical Imaging (ISBI) 2019) contains 20 3D abdominal MRI scans with total four different labels.


Liver, spleen and left and right kidney are used as semantic classes following previous settings \cite{tang2021recurrent, ding2023few, ouyang2020self}. Within each experiment, one organ is considered as unseen semantic class for testing while the rest are used for training.

\textbf{Evaluation Metrics}  We use the Dice Similarity Coefficient (DSC) to evaluate the prediction mask \textbf{m} against the ground truth mask \textbf{g}:

\begin{equation}
    \text{{DSC(m, g)}} = \frac{2 \left | m\cap g \right |}{\left | m \right |  + \left |  g \right | }
\end{equation}

\begin{figure}[ht!]
\centering
\includegraphics[width=0.3\textwidth]{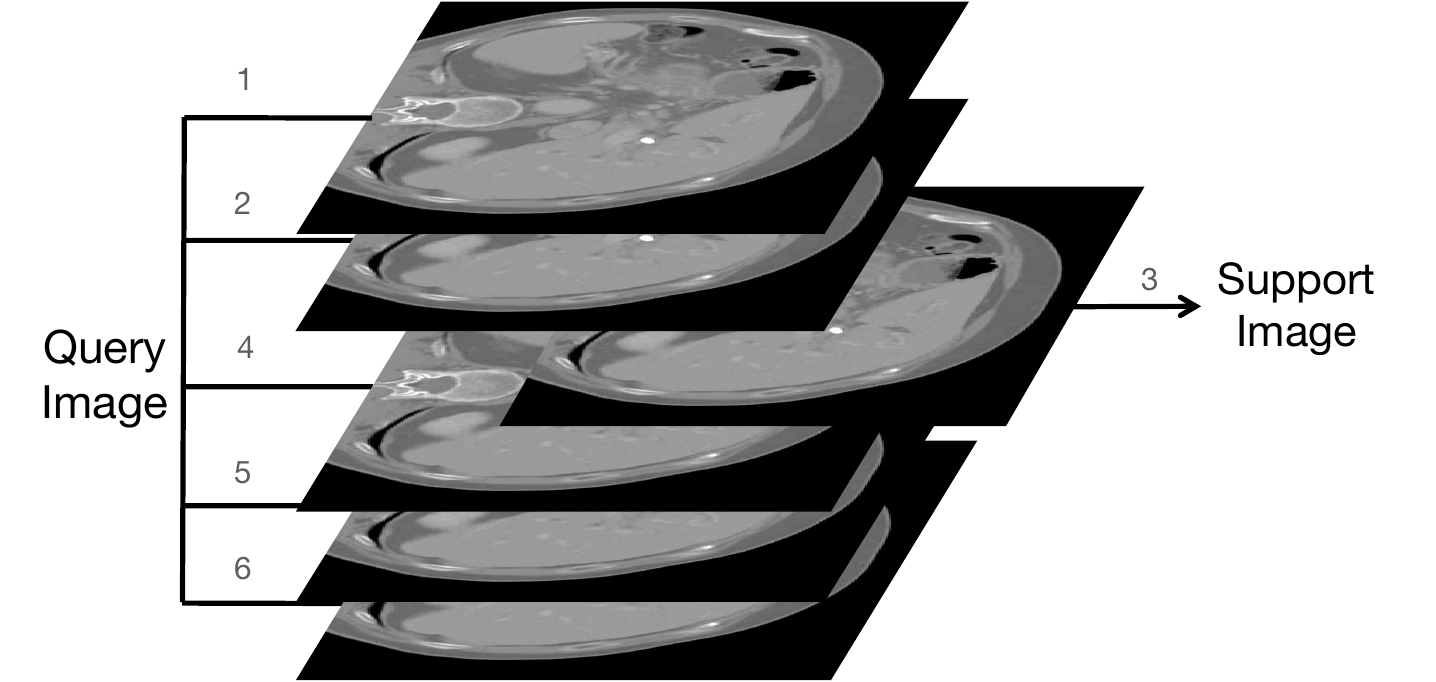}
\caption{Representative chunk}
\label{fig:4}
\end{figure}

\textbf{Implementation Details}
All the images extracted from 3D volume data are reshaped to $1024\times1024$ to fit into the SAM model. We follow the same protocol used in \cite{tang2021recurrent,ouyang2020self,roy2020squeeze} to do 1-way 1-shot learning by dividing the 3D CT scans into 12 chunks and segmenting all the query slices in one chunk by using the center slice in the chunk as the support image, see Fig.\ref{fig:4} for visual representation.
We use ViT-B \cite{dosovitskiy2021image} version of SAM and supervise our network using balanced cross entropy loss and IoU loss as \cref{loss} between the predicted mask and ground truth mask. AdamW optimizer \cite{loshchilov2017decoupled} is used for all the experiments with an initial learning rate of 2e-4. Cosine decay\cite{antoniou2018train} is applied to the learning rate. Few-shot segmentation task is trained for 50 epochs on a single NVIDIA A40 GPU and fully supervised segmentation task is trained for 50 epochs using PyTorch on a single NVIDIA RTX 3090 GPU.

\begin{figure*}[htbp]
\centering
\includegraphics[width=0.8\textwidth]{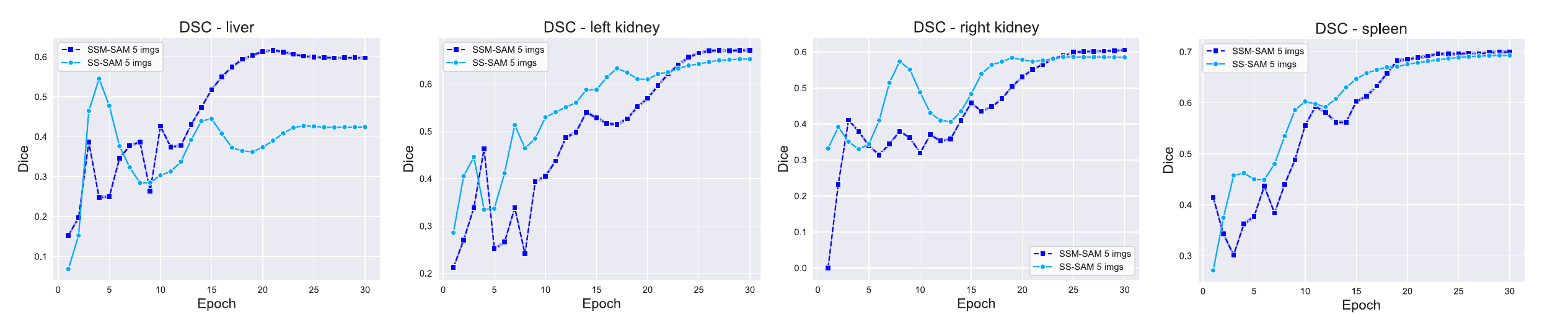}
\caption{Testing process of MAML++ based model (in deep blue) and  Fine-tuning of a model pretrained on the same distribution of tasks without MAML++ (in light blue)}
\label{fig:5}
\end{figure*}

\subsection{Main Results}   

\begin{table*}[ht!]
\centering
\begin{tabular}{|c|c|cccc|c|}
\hline Dataset & Method & Spleen $\uparrow$ & Kidney L $\uparrow$ & Kidney R $\uparrow$ & Liver $\uparrow$ & mean $\uparrow$\\
\hline
 \multirow{9}{*}{ABD-30}& SE-Net \cite{roy2020squeeze} & 0.23 & 32.83 & 14.34 & 0.27 & 11.91 \\
 &PANet \cite{wang2019panet} & 25.59 & 32.34 & 17.37 & 38.42 & 29.42 \\
 &SSL-ALPNet \cite{ouyang2020self} & 60.25 & 63.34 & 54.82 & 73.65 & 63.02 \\
 &Affine & 48.99 & 43.44 & 45.67 & 68.93 & 51.75 \\
 &RP-Net \cite{tang2021recurrent} & 72.19 & 75.03 & 70.89 & 80.53 & 74.66 \\
 \cline{2-7}
 &\textbf{SS-SAM (w/o MAML++, w/o FMAD)} & 77.24 & 72.89 & 79.78 & 72.00 & 75.48\\
 &\textbf{SS-SAM (w/o MAML++, w/ FMAD)} & 78.99 & 73.33 & 83.49 & 74.81 & 77.65\\
 &\textbf{SSM-SAM (w/ MAML++, w/o FMAD)} & 80.55 & 77.83 & 82.79 & 82.70 & 81.46\\
 \cline{2-7}
 &\textbf{SSM-SAM (w/ MAML++, w/ FMAD)} & \textbf{86.95} & \textbf{80.96} & \textbf{84.47} & \textbf{87.12} & \textbf{84.87}\\
 \hline
 \hline
 \multirow{9}{*}{ABD-MRI}& SE-Net\cite{roy2020squeeze}& 51.80 & 62.11 & 61.32 & 27.43 & 50.66\\
 & PANet\cite{wang2019panet}& 50.90 & 53.45 & 38.64 & 42.26 & 46.33\\
 & SSL-ALPNet\cite{ouyang2020self}& 67.02 & 73.63 & 78.39 & 73.05 & 73.02\\
 & Affine & 62.87 & 64.70 & 69.10 & 65.00 & 65.41\\
 & RP-Net \cite{tang2021recurrent} & 75.69 & 79.30 & \textbf{84.66} & 71.51 & 77.79\\
 \cline{2-7}
 &\textbf{SS-SAM (w/o MAML++, w/o FMAD)} & 71.99 & 76.52 & 72.13 & 69.36 & 72.50 \\
 &\textbf{SS-SAM (w/o MAML++, w/ FMAD)} & 73.64 & 78.04 & 76.69 & 72.13 & 75.12\\
 &\textbf{SSM-SAM (w/ MAML++, w/o FMAD)} & 76.77 & 80.47 & 77.44 & 76.32 & 77.75\\
 \cline{2-7}
 & \textbf{SSM-SAM (w/ MAML++, w/ FMAD)} & \textbf{78.81} & \textbf{81.70} & 80.38 & \textbf{77.50} & \textbf{79.59}\\
 \hline
\end{tabular}
\smallskip
\caption{DSC comparison with other methods on ABD-30 and ABD-MRI for few-shot learning (unit:\%).}
\label{tab:few-shot}
\end{table*}

\begin{table*}[ht!]
\centering
\begin{tabular}{|c|c|cccccccc|}
\hline 
  Method & DSC $\uparrow$ & Aorta & Gallbladder& Kidney(L) & Kidney(R) & Liver & Pancreas & Spleen & Stomach \\
\hline
 TransUNet \cite{chen2021transunet} & 77.48 & 87.23 & 63.13 & 81.87 &  77.02 & 94.08 & 55.86 & 85.08 & 75.62\\
 SwinUnet \cite{cao2022swin} & 79.13 & 85.47 & 66.53 & 83.28 & 79.61 & 94.29 & 56.58 & 90.66 & 76.60\\
 MissFormer \cite{huang2022missformer} & 81.96 & 86.99 & 68.65 & 85.21 & 82.00 & 94.41 & 65.67 & 91.92 & 80.81\\
 HiFormer \cite{heidari2023hiformer} & 80.39 & 86.21 & 65.69 & 85.23 & 79.77 & 94.61 & 59.52 & 90.99 & 81.08\\
 DAE-Former \cite{azad2022dae} & 82.43 & 88.96 & 72.30 & 86.08 & 80.88 & 94.98 & 65.12 & 91.94 & 79.19 \\
 PaNN \cite{zhou2019prior} & 90.8 & 92.5 & 72.9 & 95.3 & 92.0 & 95.3 & - & \textbf{96.8} & - \\
 \hline
 \textit{\textcolor{blue}{MissFormer}} \cite{huang2022missformer} & 
 \textit{\textcolor{blue}{78.84}} & \textit{\textcolor{blue}{85.89}} & 
 \textit{\textcolor{blue}{64.75}} & \textit{\textcolor{blue}{84.19}} &
 \textit{\textcolor{blue}{76.98}} & \textit{\textcolor{blue}{94.18}} &
 \textit{\textcolor{blue}{55.19}} & \textit{\textcolor{blue}{89.07}} & \textit{\textcolor{blue}{80.47}} \\
 
 \textit{\textcolor{blue}{HiFormer}} \cite{heidari2023hiformer} & \textit{\textcolor{blue}{78.24}} & \textit{\textcolor{blue}{85.94}} & \textit{\textcolor{blue}{60.88}} & \textit{\textcolor{blue}{84.16}} & \textit{\textcolor{blue}{76.57}} & \textit{\textcolor{blue}{94.54}} & \textit{\textcolor{blue}{54.00}} & \textit{\textcolor{blue}{89.81}} & \textit{\textcolor{blue}{80.02}} \\

 \textit{\textcolor{blue}{DAE-Former}} \cite{azad2022dae} &
 \textit{\textcolor{blue}{80.46}} & \textit{\textcolor{blue}{87.26}} & 
 \textit{\textcolor{blue}{67.46}} & \textit{\textcolor{blue}{82.15}} & \textit{\textcolor{blue}{77.53}} & \textit{\textcolor{blue}{94.36}} & \textit{\textcolor{blue}{63.32}} & \textit{\textcolor{blue}{90.85}} & \textit{\textcolor{blue}{80.77}} \\
 \hline
 SAMed \cite{zhang2023customized} & 81.88 & 87.77 & 69.11 & 80.45 & 79.95 & 94.80 & 72.17 & 88.72 & 82.06\\
 SAM Adapter \cite{chen2023sam} & 92.01 & 95.35 & 87.88 & \textbf{96.16} & 96.73 & 97.23 & \textbf{79.86} & 90.60 & 92.26 \\
\hline
 \textbf{SS-SAM} & \textbf{93.09} & \textbf{95.96} & \textbf{89.77} & 96.09 & \textbf{96.79} & \textbf{97.61} & 79.15 & 95.15 & \textbf{94.23}\\
\hline
\end{tabular}
\smallskip
\caption{Comparison to state-of-the-art models using a fully supervised method on the ABD-30 dataset. Best results are highlighted in bold (unit:\%). For a fair comparison, the results in italics are from the corresponding model with the same self-sampling module. Notably, our modified SAM (SS-SAM) outperforms other leading models.}
\label{tab:fully}
\end{table*}

\begin{table*}[ht!]
\centering
\begin{tabular}{|c|cccc|c|}
\hline 
Prompt & Spleen & Kidney L & Kidney R & Liver & Mean \\
\hline
 no prompt & 91.83 & 93.65 &  92.55 & 96.21 & 93.56 \\
 w/ position encoding & 93.05 & 95.75 & \textbf{96.87} & 97.27 & 95.74 \\
w/ self-sampling & \textbf{95.15} & \textbf{96.09} & 96.79 & \textbf{97.61} & \textbf{96.41}  \\
\hline
\end{tabular}
\smallskip
\caption{Ablation Study on no prompt, position encoding (pe) of points and self-sampling}
\label{tab:3}
\end{table*}

\textbf{Results on Few-shot Medical Image Segmentation} Table \ref{tab:few-shot} shows the few-shot segmentation performance of SSM-SAM with previous work on ABD-30 and ABD-MRI respectively. SE-Net \cite{roy2020squeeze} represents the first architecture designed explicitly for few-shot medical image segmentation. PANet \cite{wang2019panet} is an extended version of the widely-used prototypical network \cite{snell2017prototypical}, tailored for natural image segmentation. SSL-ALPNet \cite{ouyang2020self} integrates self-supervised learning with prototypical networks. Affine denotes the accuracy result after aligning the support and query images globally using an affine transformation. RP-Net \cite{tang2021recurrent} stands as the state-of-the-art framework for few-shot medical image segmentation, leveraging both a context relation encoder and a recurrent module. \cite{tang2021recurrent} reported performance for all the methods above, so these numbers are directly quoted.

First, compared to state-of-the-art method RP-Net, SSM-SAM outperforms RP-Net by 10.21\% and 1.80\% on ABD-30 and ABD-MRI respectively. Second, the use of FMAD results in improvements of 2.17\% $\sim$ 3.41\% and 1.84\% $\sim$ 2.62\% respectively. Additionally, employing the MAML++ based optimizer leads to enhancements of 5.98\% $\sim$ 7.22\% and 4.47\% $\sim$ 5.25\% respectively. This indicates that both our meta-optimizer and FMAD are highly effective.

\textbf{Results on Fully Supervised Medical Image Segmentation}

To evaluate the performance of our backbone (without MAML++), we compare SS-SAM with recent SOTA methods on ABD-30 dataset without Flexible Mask Attention Module, including U-Net \cite{ronneberger2015u}, Att-UNet \cite{oktay2018attention}, TransUnet \cite{chen2021transunet}, Swin-Unet \cite{cao2022swin}, MissFormer \cite{huang2022missformer}, TransDeepLab \cite{azad2022transdeeplab}, HiFormer \cite{heidari2023hiformer}, DAE-Former \cite{azad2022dae} and SAMed \cite{zhang2023customized} following \cite{zhang2023customized}. For a fair comparison, we integrate the self-sampling module into some of the state-of-the-art models because the original SAM and our method utilize visual prompts, which incorporate ground truth information into the model while previous state-of-the-art models did not leverage such additional information. The results are highlighted in blue italics in \cref{tab:fully}. Surprisingly, since these models are unaware of how to process this information, their performance slightly declined. We also observe that SS-SAM achieves state-of-the-art performance. These experiments demonstrate that our models are capable of achieving high performance and also refute the potential claim that introducing prompts is a form of "cheating".

\subsection{Ablation Study}  

\textbf{Effect of self-sampling.} As previously explained, the self-sampling module derives visual prompts by interpolating between the provided point coordinates and image embeddings. Subsequently, it conducts cross-attention with the image embeddings. On the other hand, SAM's point prompt employs positional encoding on the points before performing cross-attention with the image embeddings. These two approaches differ in their treatment of the sampled points. To evaluate the effectiveness of the proposed methods, we performed ablation studies on four distinct organs: the spleen, left kidney, right kidney, and liver, using a fully supervised approach. For each organ, we kept the image encoder and decoder the same, with three different prompt methods: a) no prompt, b) point prompt with position encoding, and c) point prompt with self-sampling. As presented in \cref{tab:3}, our self-sampling method achieved the highest mean

\begin{figure}[ht!]
\centering
\includegraphics[width=0.3\textwidth]{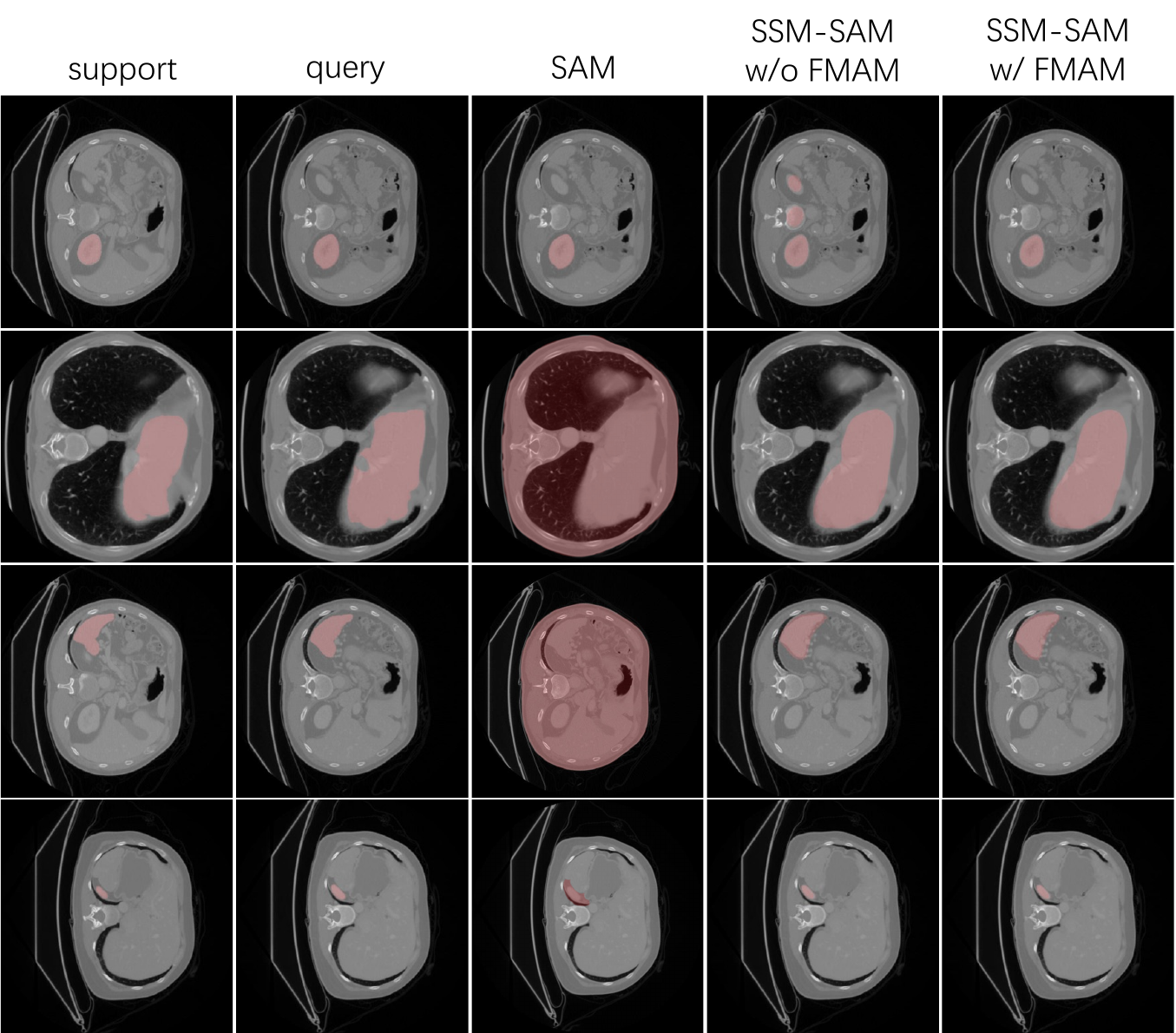}
\caption{Examples of predictions of SAM, SSM-SAM without FMAM and SSM-SAM with FMAM}
\label{result}
\end{figure}

\textbf{Effects of Meta-learner.} To highlight the few-shot learning advantages of our model achieved by integrating a meta-learner, apart from the experiment setting in \cref{tab:few-shot}, we set up a more rigorous experimental setting to compare performances with and without the meta-learner.

Following the configurations specified in \cite{finn2017model, li2020fss}, we selected only 125 images for the four organs. In each experiment, one organ is chosen as the test task (e.g., liver), while the remaining three organs (spleen, left kidney, right kidney) are used as training tasks. Only five images of each training task (spleen, left kidney, right kidney) are allocated during training. During testing, we adopt a 1-way, 5-shot approach, using five images of the test task (liver) as the support set for training. The remaining 120 images serve as the query set to evaluate the model's performance. 

Our findings were illuminating. A visual representation of the online optimization process of each of the four organs is provided in Fig.\ref{fig:5}. With just a limited number of training images, models equipped with the meta-learner outperformed those without by a significant margin of 12.07\% on average. Furthermore, models augmented by the meta-learner not only demonstrated swifter convergence in the initial epochs but also exhibited enhanced stability and performance in the concluding epochs. 

\subsection{Qualitative Results}
In Fig.\ref{result}, we present masks produced by SAM and variants of our algorithm. Notably, SAM struggles to segment smaller organs or those with indistinct boundaries. Conversely, SSM-SAM with FMAM outperforms its counterpart without FMAM, effectively minimizing the impact of similar distracting regions in the CT scans.

\section{Conclusions}
In this paper, we introduce a universal approach, SSM-SAM, designed to optimize and adapt a foundational model such as the Segment Anything Model (SAM), for few-shot medical image segmentation. Our method incorporates a positive-negative Self-Sampling prompt encoder and a Flexible Mask Attention Decoder to enhance the contextual relationship and tiny boundary information essential for mask generation. Moreover, our fast online meta-learning based optimizer facilitates high performance even without extensive training data and can be plugged into other frameworks effortlessly. Experiments demonstrate that SSM-SAM outperforms the previous state-of-the-art approach by as much as 10\% in terms of DSC. Furthermore, the proposed SSM-SAM can produce segmentations in just 50 seconds per organ, indicating its potential for real-time applications in medical settings.

{\small
\bibliographystyle{ieee_fullname}
\bibliography{egbib}

\begin{thebibliography}{10}\itemsep=-1pt

\bibitem{antoniou2018train}
Antreas Antoniou, Harrison Edwards, and Amos Storkey.
\newblock How to train your maml.
\newblock {\em arXiv preprint arXiv:1810.09502}, 2018.

\bibitem{azad2022dae}
Reza Azad, Ren{\'e} Arimond, Ehsan~Khodapanah Aghdam, Amirhosein Kazerouni, and Dorit Merhof.
\newblock Dae-former: Dual attention-guided efficient transformer for medical image segmentation.
\newblock {\em arXiv preprint arXiv:2212.13504}, 2022.

\bibitem{azad2022transdeeplab}
Reza Azad, Moein Heidari, Moein Shariatnia, Ehsan~Khodapanah Aghdam, Sanaz Karimijafarbigloo, Ehsan Adeli, and Dorit Merhof.
\newblock Transdeeplab: Convolution-free transformer-based deeplab v3+ for medical image segmentation.
\newblock In {\em International Workshop on PRedictive Intelligence In MEdicine}, pages 91--102. Springer, 2022.

\bibitem{bai2014error}
Junjie Bai and Xiaodong Wu.
\newblock Error-tolerant scribbles based interactive image segmentation.
\newblock In {\em Proceedings of the IEEE Conference on Computer Vision and Pattern Recognition}, pages 392--399, 2014.

\bibitem{batra2010icoseg}
Dhruv Batra, Adarsh Kowdle, Devi Parikh, Jiebo Luo, and Tsuhan Chen.
\newblock icoseg: Interactive co-segmentation with intelligent scribble guidance.
\newblock In {\em 2010 IEEE computer society conference on computer vision and pattern recognition}, pages 3169--3176. IEEE, 2010.

\bibitem{brown2020language}
Tom Brown, Benjamin Mann, Nick Ryder, Melanie Subbiah, Jared~D Kaplan, Prafulla Dhariwal, Arvind Neelakantan, Pranav Shyam, Girish Sastry, Amanda Askell, et~al.
\newblock Language models are few-shot learners.
\newblock {\em Advances in neural information processing systems}, 33:1877--1901, 2020.

\bibitem{cao2022swin}
Hu Cao, Yueyue Wang, Joy Chen, Dongsheng Jiang, Xiaopeng Zhang, Qi Tian, and Manning Wang.
\newblock Swin-unet: Unet-like pure transformer for medical image segmentation.
\newblock In {\em European conference on computer vision}, pages 205--218. Springer, 2022.

\bibitem{chen2021transunet}
Jieneng Chen, Yongyi Lu, Qihang Yu, Xiangde Luo, Ehsan Adeli, Yan Wang, Le Lu, Alan~L Yuille, and Yuyin Zhou.
\newblock Transunet: Transformers make strong encoders for medical image segmentation.
\newblock {\em arXiv preprint arXiv:2102.04306}, 2021.

\bibitem{chen2023sam}
Tianrun Chen, Lanyun Zhu, Chaotao Ding, Runlong Cao, Shangzhan Zhang, Yan Wang, Zejian Li, Lingyun Sun, Papa Mao, and Ying Zang.
\newblock Sam fails to segment anything?--sam-adapter: Adapting sam in underperformed scenes: Camouflage, shadow, and more.
\newblock {\em arXiv preprint arXiv:2304.09148}, 2023.

\bibitem{chen2021conditional}
Xi Chen, Zhiyan Zhao, Feiwu Yu, Yilei Zhang, and Manni Duan.
\newblock Conditional diffusion for interactive segmentation.
\newblock In {\em Proceedings of the IEEE/CVF International Conference on Computer Vision}, pages 7345--7354, 2021.

\bibitem{chen2022vision}
Zhe Chen, Yuchen Duan, Wenhai Wang, Junjun He, Tong Lu, Jifeng Dai, and Yu Qiao.
\newblock Vision transformer adapter for dense predictions.
\newblock {\em arXiv preprint arXiv:2205.08534}, 2022.

\bibitem{devlin2018bert}
Jacob Devlin, Ming-Wei Chang, Kenton Lee, and Kristina Toutanova.
\newblock Bert: Pre-training of deep bidirectional transformers for language understanding.
\newblock {\em arXiv preprint arXiv:1810.04805}, 2018.

\bibitem{ding2023few}
Hao Ding, Changchang Sun, Hao Tang, Dawen Cai, and Yan Yan.
\newblock Few-shot medical image segmentation with cycle-resemblance attention.
\newblock In {\em Proceedings of the IEEE/CVF Winter Conference on Applications of Computer Vision}, pages 2488--2497, 2023.

\bibitem{dong2018few}
Nanqing Dong and Eric~P Xing.
\newblock Few-shot semantic segmentation with prototype learning.
\newblock In {\em BMVC}, volume~3, 2018.

\bibitem{dosovitskiy2021image}
Alexey Dosovitskiy, Lucas Beyer, Alexander Kolesnikov, Dirk Weissenborn, Xiaohua Zhai, Thomas Unterthiner, Mostafa Dehghani, Matthias Minderer, Georg Heigold, Sylvain Gelly, Jakob Uszkoreit, and Neil Houlsby.
\newblock An image is worth 16x16 words: Transformers for image recognition at scale, 2021.

\bibitem{finn2017model}
Chelsea Finn, Pieter Abbeel, and Sergey Levine.
\newblock Model-agnostic meta-learning for fast adaptation of deep networks.
\newblock In {\em International conference on machine learning}, pages 1126--1135. PMLR, 2017.

\bibitem{gong20233dsamadapter}
Shizhan Gong, Yuan Zhong, Wenao Ma, Jinpeng Li, Zhao Wang, Jingyang Zhang, Pheng-Ann Heng, and Qi Dou.
\newblock 3dsam-adapter: Holistic adaptation of sam from 2d to 3d for promptable medical image segmentation, 2023.

\bibitem{he2022masked}
Kaiming He, Xinlei Chen, Saining Xie, Yanghao Li, Piotr Doll{\'a}r, and Ross Girshick.
\newblock Masked autoencoders are scalable vision learners.
\newblock In {\em Proceedings of the IEEE/CVF conference on computer vision and pattern recognition}, pages 16000--16009, 2022.

\bibitem{he2023accuracy}
Sheng He, Rina Bao, Jingpeng Li, P~Ellen Grant, and Yangming Ou.
\newblock Accuracy of segment-anything model (sam) in medical image segmentation tasks.
\newblock {\em arXiv preprint arXiv:2304.09324}, 2023.

\bibitem{heidari2023hiformer}
Moein Heidari, Amirhossein Kazerouni, Milad Soltany, Reza Azad, Ehsan~Khodapanah Aghdam, Julien Cohen-Adad, and Dorit Merhof.
\newblock Hiformer: Hierarchical multi-scale representations using transformers for medical image segmentation.
\newblock In {\em Proceedings of the IEEE/CVF Winter Conference on Applications of Computer Vision}, pages 6202--6212, 2023.

\bibitem{houlsby2019parameter}
Neil Houlsby, Andrei Giurgiu, Stanislaw Jastrzebski, Bruna Morrone, Quentin De~Laroussilhe, Andrea Gesmundo, Mona Attariyan, and Sylvain Gelly.
\newblock Parameter-efficient transfer learning for nlp.
\newblock In {\em International Conference on Machine Learning}, pages 2790--2799. PMLR, 2019.

\bibitem{hu2023skinsam}
Mingzhe Hu, Yuheng Li, and Xiaofeng Yang.
\newblock Skinsam: Empowering skin cancer segmentation with segment anything model.
\newblock {\em arXiv preprint arXiv:2304.13973}, 2023.

\bibitem{huang2022missformer}
Xiaohong Huang, Zhifang Deng, Dandan Li, Xueguang Yuan, and Ying Fu.
\newblock Missformer: An effective transformer for 2d medical image segmentation.
\newblock {\em IEEE Transactions on Medical Imaging}, 2022.

\bibitem{jang2019interactive}
Won-Dong Jang and Chang-Su Kim.
\newblock Interactive image segmentation via backpropagating refinement scheme.
\newblock In {\em Proceedings of the IEEE/CVF Conference on Computer Vision and Pattern Recognition}, pages 5297--5306, 2019.

\bibitem{ji2023segment}
Wei Ji, Jingjing Li, Qi Bi, Tingwei Liu, Wenbo Li, and Li Cheng.
\newblock Segment anything is not always perfect: An investigation of sam on different real-world applications, 2023.

\bibitem{kavur2021chaos}
A~Emre Kavur, N~Sinem Gezer, Mustafa Bar{\i}{\c{s}}, Sinem Aslan, Pierre-Henri Conze, Vladimir Groza, Duc~Duy Pham, Soumick Chatterjee, Philipp Ernst, Sava{\c{s}} {\"O}zkan, et~al.
\newblock Chaos challenge-combined (ct-mr) healthy abdominal organ segmentation.
\newblock {\em Medical Image Analysis}, 69:101950, 2021.

\bibitem{kirillov2023segment}
Alexander Kirillov, Eric Mintun, Nikhila Ravi, Hanzi Mao, Chloe Rolland, Laura Gustafson, Tete Xiao, Spencer Whitehead, Alexander~C. Berg, Wan-Yen Lo, Piotr Dollár, and Ross Girshick.
\newblock Segment anything, 2023.

\bibitem{MICCAI}
B Landman, Z Xu, J Igelsias, M Styner, T Langerak, and A Klein.
\newblock Miccai multi-atlas labeling beyond the cranial vault–workshop and challenge.
\newblock {\em Proc. MICCAI Multi-Atlas Labeling Beyond Cranial Vault—Workshop Challenge}, 2015.

\bibitem{li2021adaptive}
Gen Li, Varun Jampani, Laura Sevilla-Lara, Deqing Sun, Jonghyun Kim, and Joongkyu Kim.
\newblock Adaptive prototype learning and allocation for few-shot segmentation.
\newblock In {\em Proceedings of the IEEE/CVF conference on computer vision and pattern recognition}, pages 8334--8343, 2021.

\bibitem{li2020fss}
Xiang Li, Tianhan Wei, Yau~Pun Chen, Yu-Wing Tai, and Chi-Keung Tang.
\newblock Fss-1000: A 1000-class dataset for few-shot segmentation.
\newblock In {\em Proceedings of the IEEE/CVF conference on computer vision and pattern recognition}, pages 2869--2878, 2020.

\bibitem{li2023polyp}
Yuheng Li, Mingzhe Hu, and Xiaofeng Yang.
\newblock Polyp-sam: Transfer sam for polyp segmentation.
\newblock {\em arXiv preprint arXiv:2305.00293}, 2023.

\bibitem{li2022exploring}
Yanghao Li, Hanzi Mao, Ross Girshick, and Kaiming He.
\newblock Exploring plain vision transformer backbones for object detection.
\newblock In {\em European Conference on Computer Vision}, pages 280--296. Springer, 2022.

\bibitem{lin2016scribblesup}
Di Lin, Jifeng Dai, Jiaya Jia, Kaiming He, and Jian Sun.
\newblock Scribblesup: Scribble-supervised convolutional networks for semantic segmentation.
\newblock In {\em Proceedings of the IEEE conference on computer vision and pattern recognition}, pages 3159--3167, 2016.

\bibitem{lin2020interactive}
Zheng Lin, Zhao Zhang, Lin-Zhuo Chen, Ming-Ming Cheng, and Shao-Ping Lu.
\newblock Interactive image segmentation with first click attention.
\newblock In {\em Proceedings of the IEEE/CVF conference on computer vision and pattern recognition}, pages 13339--13348, 2020.

\bibitem{liu2022simpleclick}
Qin Liu, Zhenlin Xu, Gedas Bertasius, and Marc Niethammer.
\newblock Simpleclick: Interactive image segmentation with simple vision transformers.
\newblock {\em arXiv preprint arXiv:2210.11006}, 2022.

\bibitem{liu2023explicit}
Weihuang Liu, Xi Shen, Chi-Man Pun, and Xiaodong Cun.
\newblock Explicit visual prompting for low-level structure segmentations.
\newblock In {\em Proceedings of the IEEE/CVF Conference on Computer Vision and Pattern Recognition}, pages 19434--19445, 2023.

\bibitem{loshchilov2017decoupled}
Ilya Loshchilov and Frank Hutter.
\newblock Decoupled weight decay regularization.
\newblock {\em arXiv preprint arXiv:1711.05101}, 2017.

\bibitem{ma2023segment}
Jun Ma and Bo Wang.
\newblock Segment anything in medical images.
\newblock {\em arXiv preprint arXiv:2304.12306}, 2023.

\bibitem{makarevich2021metamedseg}
Anastasia Makarevich, Azade Farshad, Vasileios Belagiannis, and Nassir Navab.
\newblock Metamedseg: Volumetric meta-learning for few-shot organ segmentation, 2021.

\bibitem{mazurowski2023segment}
Maciej~A Mazurowski, Haoyu Dong, Hanxue Gu, Jichen Yang, Nicholas Konz, and Yixin Zhang.
\newblock Segment anything model for medical image analysis: an experimental study.
\newblock {\em Medical Image Analysis}, page 102918, 2023.

\bibitem{oktay2018attention}
Ozan Oktay, Jo Schlemper, Loic~Le Folgoc, Matthew Lee, Mattias Heinrich, Kazunari Misawa, Kensaku Mori, Steven McDonagh, Nils~Y Hammerla, Bernhard Kainz, et~al.
\newblock Attention u-net: Learning where to look for the pancreas.
\newblock {\em arXiv preprint arXiv:1804.03999}, 2018.

\bibitem{ouyang2020self}
Cheng Ouyang, Carlo Biffi, Chen Chen, Turkay Kart, Huaqi Qiu, and Daniel Rueckert.
\newblock Self-supervision with superpixels: Training few-shot medical image segmentation without annotation.
\newblock In {\em Computer Vision--ECCV 2020: 16th European Conference, Glasgow, UK, August 23--28, 2020, Proceedings, Part XXIX 16}, pages 762--780. Springer, 2020.

\bibitem{rajchl2016deepcut}
Martin Rajchl, Matthew~CH Lee, Ozan Oktay, Konstantinos Kamnitsas, Jonathan Passerat-Palmbach, Wenjia Bai, Mellisa Damodaram, Mary~A Rutherford, Joseph~V Hajnal, Bernhard Kainz, et~al.
\newblock Deepcut: Object segmentation from bounding box annotations using convolutional neural networks.
\newblock {\em IEEE transactions on medical imaging}, 36(2):674--683, 2016.

\bibitem{ronneberger2015u}
Olaf Ronneberger, Philipp Fischer, and Thomas Brox.
\newblock U-net: Convolutional networks for biomedical image segmentation.
\newblock In {\em Medical Image Computing and Computer-Assisted Intervention--MICCAI 2015: 18th International Conference, Munich, Germany, October 5-9, 2015, Proceedings, Part III 18}, pages 234--241. Springer, 2015.

\bibitem{roy2020squeeze}
Abhijit~Guha Roy, Shayan Siddiqui, Sebastian P{\"o}lsterl, Nassir Navab, and Christian Wachinger.
\newblock ‘squeeze \& excite’guided few-shot segmentation of volumetric images.
\newblock {\em Medical image analysis}, 59:101587, 2020.

\bibitem{roy2023sam}
Saikat Roy, Tassilo Wald, Gregor Koehler, Maximilian~R Rokuss, Nico Disch, Julius Holzschuh, David Zimmerer, and Klaus~H Maier-Hein.
\newblock Sam. md: Zero-shot medical image segmentation capabilities of the segment anything model.
\newblock {\em arXiv preprint arXiv:2304.05396}, 2023.

\bibitem{shi2023generalist}
Peilun Shi, Jianing Qiu, Sai Mu~Dalike Abaxi, Hao Wei, Frank P-W Lo, and Wu Yuan.
\newblock Generalist vision foundation models for medical imaging: A case study of segment anything model on zero-shot medical segmentation.
\newblock {\em Diagnostics}, 13(11):1947, 2023.

\bibitem{singh2021metamed}
Rishav Singh, Vandana Bharti, Vishal Purohit, Abhinav Kumar, Amit~Kumar Singh, and Sanjay~Kumar Singh.
\newblock Metamed: Few-shot medical image classification using gradient-based meta-learning.
\newblock {\em Pattern Recognition}, 120:108111, 2021.

\bibitem{snell2017prototypical}
Jake Snell, Kevin Swersky, and Richard Zemel.
\newblock Prototypical networks for few-shot learning.
\newblock {\em Advances in neural information processing systems}, 30, 2017.

\bibitem{tang2021recurrent}
Hao Tang, Xingwei Liu, Shanlin Sun, Xiangyi Yan, and Xiaohui Xie.
\newblock Recurrent mask refinement for few-shot medical image segmentation.
\newblock In {\em Proceedings of the IEEE/CVF international conference on computer vision}, pages 3918--3928, 2021.

\bibitem{vaswani2017attention}
Ashish Vaswani, Noam Shazeer, Niki Parmar, Jakob Uszkoreit, Llion Jones, Aidan~N Gomez, {\L}ukasz Kaiser, and Illia Polosukhin.
\newblock Attention is all you need.
\newblock {\em Advances in neural information processing systems}, 30, 2017.

\bibitem{wang2018interactive}
Guotai Wang, Wenqi Li, Maria~A Zuluaga, Rosalind Pratt, Premal~A Patel, Michael Aertsen, Tom Doel, Anna~L David, Jan Deprest, S{\'e}bastien Ourselin, et~al.
\newblock Interactive medical image segmentation using deep learning with image-specific fine tuning.
\newblock {\em IEEE transactions on medical imaging}, 37(7):1562--1573, 2018.

\bibitem{wang2023review}
Jiaqi Wang, Zhengliang Liu, Lin Zhao, Zihao Wu, Chong Ma, Sigang Yu, Haixing Dai, Qiushi Yang, Yiheng Liu, Songyao Zhang, Enze Shi, Yi Pan, Tuo Zhang, Dajiang Zhu, Xiang Li, Xi Jiang, Bao Ge, Yixuan Yuan, Dinggang Shen, Tianming Liu, and Shu Zhang.
\newblock Review of large vision models and visual prompt engineering, 2023.

\bibitem{wang2019panet}
Kaixin Wang, Jun~Hao Liew, Yingtian Zou, Daquan Zhou, and Jiashi Feng.
\newblock Panet: Few-shot image semantic segmentation with prototype alignment.
\newblock In {\em proceedings of the IEEE/CVF international conference on computer vision}, pages 9197--9206, 2019.

\bibitem{wang2021transformer}
Ning Wang, Wengang Zhou, Jie Wang, and Houqiang Li.
\newblock Transformer meets tracker: Exploiting temporal context for robust visual tracking.
\newblock In {\em Proceedings of the IEEE/CVF conference on computer vision and pattern recognition}, pages 1571--1580, 2021.

\bibitem{wu2023medical}
Junde Wu, Rao Fu, Huihui Fang, Yuanpei Liu, Zhaowei Wang, Yanwu Xu, Yueming Jin, and Tal Arbel.
\newblock Medical sam adapter: Adapting segment anything model for medical image segmentation.
\newblock {\em arXiv preprint arXiv:2304.12620}, 2023.

\bibitem{wu2014milcut}
Jiajun Wu, Yibiao Zhao, Jun-Yan Zhu, Siwei Luo, and Zhuowen Tu.
\newblock Milcut: A sweeping line multiple instance learning paradigm for interactive image segmentation.
\newblock In {\em Proceedings of the IEEE conference on computer vision and pattern recognition}, pages 256--263, 2014.

\bibitem{xu2016deep}
Ning Xu, Brian Price, Scott Cohen, Jimei Yang, and Thomas~S Huang.
\newblock Deep interactive object selection.
\newblock In {\em Proceedings of the IEEE conference on computer vision and pattern recognition}, pages 373--381, 2016.

\bibitem{yu2020foal}
Hanchao Yu, Shanhui Sun, Haichao Yu, Xiao Chen, Honghui Shi, Thomas~S Huang, and Terrence Chen.
\newblock Foal: Fast online adaptive learning for cardiac motion estimation.
\newblock In {\em Proceedings of the IEEE/CVF conference on computer vision and pattern recognition}, pages 4313--4323, 2020.

\bibitem{zhang2021self}
Bingfeng Zhang, Jimin Xiao, and Terry Qin.
\newblock Self-guided and cross-guided learning for few-shot segmentation.
\newblock In {\em Proceedings of the IEEE/CVF Conference on Computer Vision and Pattern Recognition}, pages 8312--8321, 2021.

\bibitem{zhang2023survey}
Chaoning Zhang, Sheng Zheng, Chenghao Li, Yu Qiao, Taegoo Kang, Xinru Shan, Chenshuang Zhang, Caiyan Qin, Francois Rameau, Sung-Ho Bae, et~al.
\newblock A survey on segment anything model (sam): Vision foundation model meets prompt engineering.
\newblock {\em arXiv preprint arXiv:2306.06211}, 2023.

\bibitem{zhang2023customized}
Kaidong Zhang and Dong Liu.
\newblock Customized segment anything model for medical image segmentation.
\newblock {\em arXiv preprint arXiv:2304.13785}, 2023.

\bibitem{zhou2023sam}
Tao Zhou, Yizhe Zhang, Yi Zhou, Ye Wu, and Chen Gong.
\newblock Can sam segment polyps?, 2023.

\bibitem{zhou2019prior}
Yuyin Zhou, Zhe Li, Song Bai, Chong Wang, Xinlei Chen, Mei Han, Elliot Fishman, and Alan~L Yuille.
\newblock Prior-aware neural network for partially-supervised multi-organ segmentation.
\newblock In {\em Proceedings of the IEEE/CVF international conference on computer vision}, pages 10672--10681, 2019.

\bibitem{zou2023segment}
Xueyan Zou, Jianwei Yang, Hao Zhang, Feng Li, Linjie Li, Jianfeng Gao, and Yong~Jae Lee.
\newblock Segment everything everywhere all at once.
\newblock {\em arXiv preprint arXiv:2304.06718}, 2023.

\end{thebibliography}
}

\end{document}